\documentclass[letterpaper, 10 pt, conference]{ieeeconf}  

\IEEEoverridecommandlockouts                              

\usepackage{graphics} 
\usepackage{epsfig} 
\usepackage{mathptmx} 
\usepackage{times} 
\usepackage{amsmath,amsfonts}
\usepackage{algorithmic}
\usepackage{array}
\usepackage[caption=false,font=normalsize,labelfont=sf,textfont=sf]{subfig}
\usepackage{textcomp}
\usepackage{stfloats}
\usepackage{url}
\usepackage{verbatim}
\usepackage{graphicx}
\def\BibTeX{{\rm B\kern-.05em{\sc i\kern-.025em b}\kern-.08em
    T\kern-.1667em\lower.7ex\hbox{E}\kern-.125emX}}
\usepackage{balance}
\usepackage{multirow}
\usepackage{tabularx}
\usepackage{booktabs}
\usepackage{cite}
\usepackage{xcolor}
\usepackage{array}
\usepackage{colortbl}
\usepackage{soul}

\definecolor{linkcolor}{RGB}{255,0,0}
\definecolor{urlcolor}{RGB}{255,105,180}
\definecolor{citecolor}{RGB}{66,168,235}
\usepackage[pagebackref=true,breaklinks=true,letterpaper=true,colorlinks,bookmarks=false]{hyperref}
\hypersetup{colorlinks=true,linkcolor=linkcolor,urlcolor=urlcolor,citecolor=blue}

\title{\LARGE \bf
VG4D: Vision-Language Model Goes 4D Video Recognition
}

\author{Zhichao Deng$^{1}$, Xiangtai Li$^{2}$, Xia Li$^{3}$, Yunhai Tong$^{2}$, Shen Zhao$^{1*}$, and Mengyuan Liu$^{4*}$ 
\thanks{*Corresponding author: Mengyuan Liu and Shen Zhao.}
\thanks{$^{1}$ Zhichao Deng and Shen Zhao are with the School of Intelligent Systems Engineering, Sun Yat-sen University, Shenzhen, China{}
        {\tt\small dengzhch3@mail2.sysu.edu.cn, z-s-06@163.com)}}%
\thanks{$^{2}$ Xiangtai Li and Yunhai Tong are with the National Key Laboratory of General Artificial Intelligence, Peking University, and Tong is also with PKU-Wuhan Institute for Artificial Intelligence.
        {\tt\small lxtpku@pku.edu.cn}}%
\thanks{$^{3}$ Xia Li is with the Department of Computer Science, ETH Zurich.
        {\tt\small xia.li@inf.ethz.ch}}%
\thanks{$^{4}$ Mengyuan Liu is with the National Key Laboratory of General Artificial Intelligence, Peking University, Shenzhen Graduate School
        {\tt\small liumengyuan@pku.edu.cn}}%
\thanks{This work was supported by the National Natural Science Foundation of China (No. 62203476), the Natural Science Foundation of Shenzhen (No. JCYJ20230807120801002), the National Key Research and Development Program of China (No. 2023YFC3807600), the interdisciplinary doctoral grants (iDoc 2021-360) from the Personalized Health and Related Technologies (PHRT) of the ETH domain, Switzerland.}
}

\begin{document}

\maketitle

\begin{abstract}
    Understanding the real world through point cloud video is a crucial aspect of robotics and autonomous driving systems. 
    However, prevailing methods for 4D point cloud recognition
    have limitations due to sensor resolution, which leads to a lack of detailed information.
    Recent advances have shown that Vision-Language Models (VLM) pre-trained on web-scale text-image datasets can learn fine-grained visual concepts that can be transferred to various downstream tasks. 
    However, effectively integrating VLM into the domain of 4D point clouds remains an unresolved problem. 
    In this work, we propose the Vision-Language Models Goes 4D (VG4D) framework to transfer VLM knowledge from visual-text pre-trained models to a 4D point cloud network. 
    Our approach involves aligning the 4D encoder's representation with a VLM to learn a shared visual and text space from training on large-scale image-text pairs. By transferring the knowledge of the VLM to the 4D encoder and combining the VLM, our VG4D achieves improved recognition performance. 
    To enhance the 4D encoder, we modernize the classic dynamic point cloud backbone and propose an improved version of PSTNet, im-PSTNet, which can efficiently model point cloud videos. 
    Experiments demonstrate that our method achieves state-of-the-art performance for action recognition on both the NTU RGB+D 60 dataset and the NTU RGB+D 120 dataset.
    Code is available at \url{https://github.com/Shark0-0/VG4D}.
\end{abstract}


\section{Introduction}






Recently, robotics and autonomous driving systems have used real-time depth sensors such as LiDARs to achieve 3D perception~\cite{ICRApc1,ICRArgb2,huang2023visual,chen2023open}. Point clouds from LiDARs 
can provide rich geometric information and facilitate the machine's comprehension of environmental perception. 
Early methods~\cite{ral9961854,DBLP:conf/nips/QiYSG17,fang2024explore,fang2024modelnet} focus on parsing the real world from static point clouds, neglecting temporal changes. 
To better understand the time-varying world, recent research focuses more on understanding point cloud videos in 4D, encompassing three spatial dimensions and one temporal dimension.
Several works~\cite{ovsep20204d,DBLP:journals/pami/FanYK23} have been done in 4D point cloud modeling. These methods either aim to design improved networks for modeling 4D point clouds or employ self-supervised methods to enhance the efficiency of 4D point cloud representations. 

\begin{figure}[!t]
\centering
\includegraphics[width=3.5in]{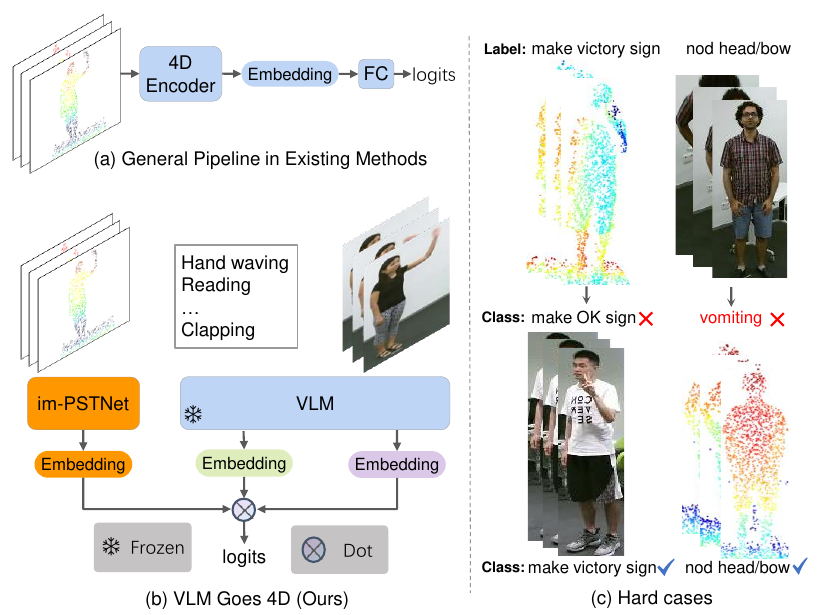}
\caption{(a) General Pipeline in Existing Methods: Input point cloud video is processed by a 4D encoder and then a standard classifier to generate prediction scores. (b) Our proposed method harnesses the knowledge of a Visual-Language pre-trained model to enhance action recognition performance. (c) Some classification hard cases of point cloud and RGB.}
\vspace{-1em}
\label{fig1}
\end{figure}

Nonetheless, the recognition of 4D videos remains a challenge for machines, primarily due to the inherent characteristics of point cloud data, which lack texture information. Owing to the lower resolution of LiDARs, point clouds may lack some details, resulting in the loss of fine-grained information.
The failure recognition case of the traditional 4D point cloud network is shown on the right side of Fig.\ref{fig1}. We find that recognition failures of 4D point clouds occur due to the small differences between actions.
%
When a person executes a specific action with limbs, the limb engaged and the object being manipulated are vital in distinguishing the action. 
For example, when differentiating between actions such as ``making a victory sign" and ``making an OK sign," the texture characteristics of the hand become critical for precise recognition of the respective actions. On the other hand, images from the RGB modality lack depth information. 
When a person in the video moves deeply, changes in depth become challenging to discern, leading to confusion between actions involving depth movement, such as ``nod bow" and others like ``vomiting."
\textbf{In conclusion, point clouds typically struggle to provide fine-grained information effectively, and it is difficult for RGB to provide depth information.} 

 \textbf{Conversely, the rich texture of RGB inherently provides an abundance of fine-grained information, whereas the presence of depth information is inherent to the point cloud modality.} Recently, Vision-Language Models (VLM), pre-trained on large-scale image-text pairs, were adept at learning fine-grained visual concepts from RGB images. Some researches~\cite{Zeng_2023_CVPR,hegde2023clip,hanoonavificlip, wu2023open, xu2023dst} have successfully applied the knowledge gleaned from VLMs to 3D static point cloud and video understanding. However, due to the modality gap,
RGB video and 4D point cloud understanding are treated as two distinct problems. 
The RGB video and 4D point cloud models are constructed with disparate architectures and trained on varying data types. \textbf{Currently, the utilization of RGB video models to recognize 4D point clouds has not been extensively explored.}

\textbf{Motivated by VLM for point processing and video understanding}~\cite{Xue_2023_CVPR, XCLIP}, we make the first study of VLM in 4D action understanding within this work.
Specifically, we propose a new VG4D (VLM Goes 4D) framework that trains a 4D encoder by leveraging VLM's knowledge. 
To jointly fine-tune this 4D encoder, we put forth a cross-modal contrastive learning approach, which facilitates harmonious alignment of the 3D representation with the VLM feature domain.
In this manner, the deficiency in fine-grained details within the point cloud modality is effectively compensated by rich content information in the RGB modality. Building on the foundation of VG4D, we synergize the exceptional capabilities of Vision-Language Models (VLMs) in video understanding with 4D point cloud representation to enhance multi-modal action recognition. Specifically, as shown in Fig.~\ref{fig1}(b), after aligning across multiple modalities, we achieve robust multi-modal action recognition by integrating multi-modal prediction scores and utilizing text information as classifiers. In addition, most VLMs are open source and easy to obtain, and we apply different VLMs integrated into our proposed VG4D framework, demonstrating the generalization of our proposed method.

Moreover, we revisit PSTNet, the classical and widely used network in dynamic point cloud modeling.
We observe that a significant portion of the performance improvement achieved by the state-of-the-art approach compared to PSTNet can be attributed to the enhanced training strategy employed. Through comprehensive experimentation with these enhancements, we identified pivotal elements contributing to variations in performance outcomes. Subsequently, we employed these identified components to revitalize the architecture of PSTNet. As a result, we propose the improved version of PSTNet dubbed im-PSTNet, 
and use it as the 4D encoder for the VG4D framework.
Our contributions can be summarized as follows:


\begin{itemize}
  \item We propose a novel VG4D framework that aligns 4D feature space with language-image representation, facilitating point cloud video recognition. To our knowledge, we are the first to explore transferring VLM's pre-trained knowledge to 4D point cloud understanding.
  \item We design a 4D encoder, named im-PSTNet, as the encoder of point cloud video, which achieves an effective point cloud video representation. We combine the robust characterization capabilities of VLMs with the outstanding point cloud video understanding abilities of the 4D network to enhance multi-modal action recognition. 
  \item Our proposed VG4D significantly outperforms the recent state-of-the-art methods on large-scale RGB+D action recognition datasets.
\end{itemize}

\section{RELATED WORK}

\noindent
\textbf{Static Point Cloud Processing.} 
Much progress~\cite{ral9956014} has been made in the field of static point cloud analysis. Point-based methods directly process raw point clouds as input. 
They are pioneered by PointNet~\cite{DBLP:conf/cvpr/QiSMG17}, which models the permutation invariance of points with shared MLPs by point-wised feature extraction. 
PointNet++~\cite{DBLP:conf/nips/QiYSG17} enhances PointNet by capturing local geometric structures. Later on, point-based methods~\cite{qian2022pointnext,dgcnn,fang2024modelnet} aim at designing local aggregation operators for local feature extraction. 
Recently, several methods have leveraged Transformer-like networks~\cite{DBLP:conf/iccv/ZhaoJJTK21, DBLP:journals/ral/ChenWKYR23} to extract information via self-attention. 
Meanwhile, self-supervised representation learning methods~\cite{DBLP:conf/cvpr/YuTR00L22,DBLP:conf/eccv/PangWTLTY22,yan2022let,fang2023explore} such as contrastive learning and reconstruction masks have attracted significant interest from the community. 
However, these methods mainly focus on static point clouds and cannot directly process dynamic point cloud video due to the lack of the temporal dynamics of point clouds into account. 

\noindent
\textbf{4D Point Cloud Modeling.}
Dynamic 4D point cloud modeling is more challenging than static point cloud processing. 
Previous point cloud video recognition methods rely on \textit{convolutional}, \textit{multi-stream}, or \textit{vision transformer}. 
Within a convolutional framework, MeteorNet~\cite{DBLP:conf/iccv/LiuYB19} is the first method on deep learning for dynamic raw 4D point cloud, which extends 3D points to 4D points and then appends a temporal dimension to PointNet++ to process these 4D points. 
Meanwhile, PSTNet~\cite{fan2021pstnet,DBLP:journals/pami/FanYYK22} models spatio-temporal information of raw point cloud videos via decomposing space and time hierarchically.
For multi-stream design, the representative method 3DV~\cite{DBLP:conf/cvpr/YanchengWang} integrates 3D motion information into a regular compact voxel set and then applies PointNet++ to extract representations from the set for 3D action recognition. 
Some other methods~\cite{DBLP:conf/cvpr/ZhongZH0TM22,tcsv/GMNet,tmm/GMTNet,wang2024gcnext,wang2023skeleton,wang2023dynamic} decouple spatial and temporal information. 
Recently, P4Transformer~\cite{fan21p4transformer} and other methods~\cite{DBLP:conf/eccv/WenLHDY22,DBLP:conf/wacv/WeiLXKG22} adopt self-attention module to capture the long-range spatial-temporal context information.
In addition, recent methods~\cite{DBLP:conf/mm/ChenLL0H022,Shen_2023_CVPR} start exploring the application of self-supervised or semi-supervised learning in dynamic 4D point cloud modeling. 
However, owing to the absence of fine-grained appearance input, the aforementioned methods still encounter difficulties in recognition.

\noindent
\textbf{Vision-Language Models.} Large vision-language models, comprising an image and a text encoder, are trained on extensive image-text pairs in a contrastive manner to learn a shared feature space between images and textual labels. 
For example, CLIP is transferable to a variety of downstream tasks, including point cloud understanding~\cite{hegde2023clip, Zeng_2023_CVPR}, video understanding~\cite{huang2022clip2point,yang2023aim,bike}, etc. 
Recently, several studies have extended the existing CLIP model to the video domain. AIM~\cite{yang2023aim} reuses the CLIP self-attention as the temporal ones via an additional adapter module. 
Vita-CLIP~\cite{wasim2023vitaclip} proposed a multi-modal prompt learning scheme to balance the supervised and zero-shot performance.
%
%
However, unlike the video domain, 3D point cloud processing with VLMs is still in its infancy. 
PointCLIP~\cite{DBLP:conf/cvpr/ZhangGZLM0QG022} directly uses the depth maps of 3D point clouds as the input of CLIP to perform zero-shot classification. 
ULIP~\cite{Xue_2023_CVPR} learns a unified representation among image, text, and point cloud that enhances 3D representation learning. 
Our method is different from PointCLIP and ULIP. 
In particular, we aim to use the fine-grained features learned by VLM to improve 4D point cloud recognition, which can compensate for the shortcomings of missing details in 3D point clouds.

\section{Method}

%

\begin{figure*}[!t]
\centering
\includegraphics[width=\textwidth]{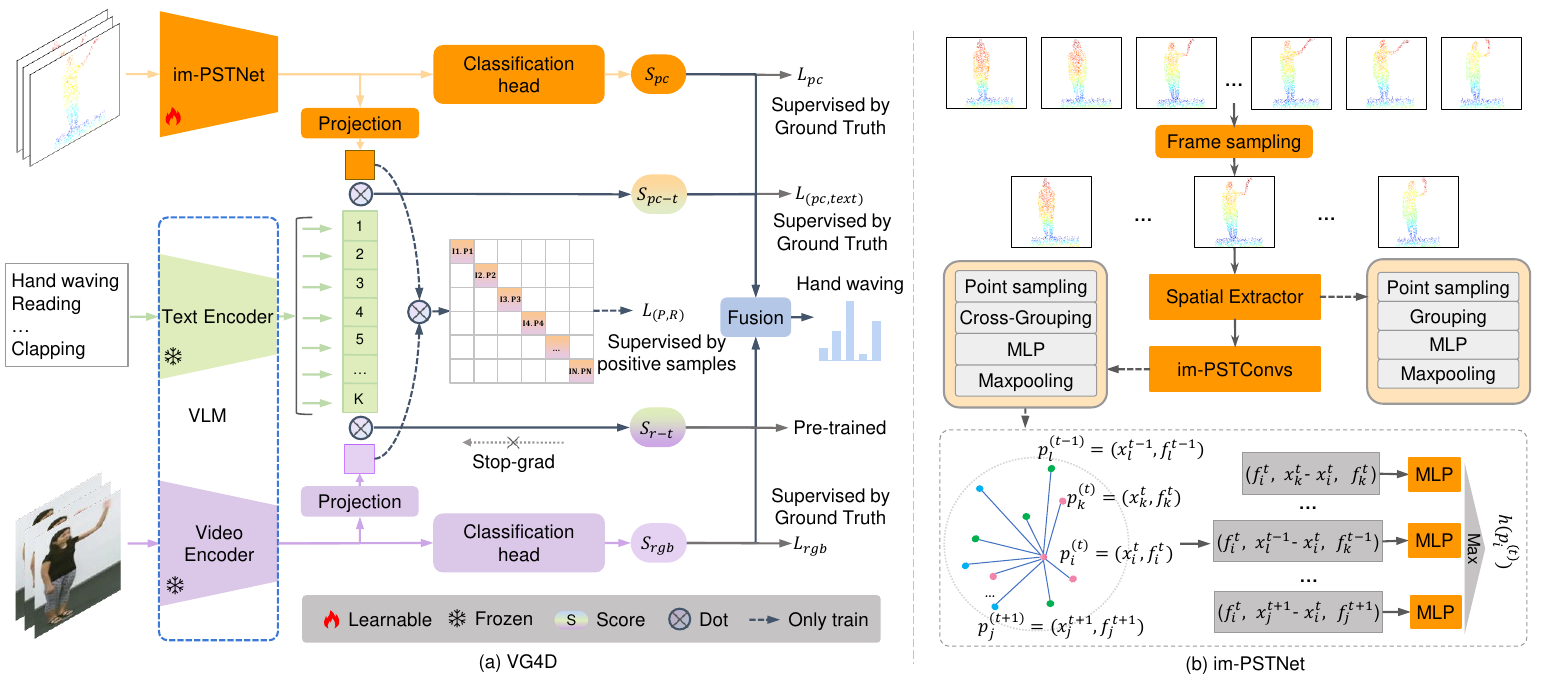}
\caption{Overall architecture of our framework. (a) VG4D (VLM goes 4D). We use a cross-modal contrastive learning objective to train our proposed 4D encoder: im-PSTNet. The knowledge of the VLM is transferred to the 4D encoder by aligning the 4D representation with language and RGB, respectively. During testing, an ensemble approach is used to integrate multiple scores. (b) The overall framework of our proposed im-PSTNet. It consists of a spatial feature extractor and a spatio-temporal feature extractor.}
\label{fig2}   
\end{figure*}

In this section, we first give an overview of the proposed VG4D framework, followed by details on how to train the network effectively. 
Then, we describe the design of im-PSTNet, which is an improved 4D encoder proposed for VG4D.

\subsection{Overview of VG4D}
As illustrated in Fig.~\ref{fig2}, our proposed VG4D framework consists of 3 networks: 4D point cloud encoder $E _ { P }$, video encoder $E _ { V }$ and text encoder $E _ { T }$ from VLM. We use language-RGB-4D point cloud triplets to train the framework. The RGB video and 4D point cloud are obtained from a camera and depth sensor, which captures data from the same sample. 
In addition, the language component consists of textual descriptions of all possible action categories. We define a set of N language-RGB video-4D point cloud triplets as $\{ x _ { i } ^ { P }, x _ { i } ^ { V }, x _ { i } ^ { T }\} _ { i = 1 } ^ { N }$, where $x _ { i } ^ { P }$ represents 4D point cloud, $x _ { i } ^ { V }$ is the RGB video, and $x _ { i } ^ { T }$ is the text of action category.

\noindent
\textbf{4D encoder} takes in a point cloud video as the input.  The output of the 4D encoder is a feature vector that encapsulates motion details. To maintain congruence in feature dimensions with other modalities, we introduced an additional projection layer for the 4D point cloud output. Given a 4D point cloud $x _ { i } ^ { P }$, we feed the point cloud video $x _ { i } ^ { P }$ into the 4D encoder $E _ { P }$ to obtain the feature representation in a common embedding space by projecting the encoded feature to a common dimension represented by: 
\begin{equation}
f_i^{P} = \psi _ { P } ( E _ { P } ( x _ { i } ^ { P } ))
\end{equation}
where $ f_i^{P} \in \mathbb{R} ^ { C _ { P } }$ represent the 4D point cloud embedding after the projection layers $\psi _ { P }$. To better learn action information from the 4D point cloud, we incorporate our custom-designed im-PSTNet as the 4D encoder. A comprehensive explanation of the im-PSTNet's particulars will be presented in Sect.\ref{im-PSTNet}.

\noindent
\textbf{VLM} takes in the RGB video and action category texts as the input. In video understanding, VLMs learn to synchronize video representations with their corresponding textual counterparts through the collaborative training of a video and text encoder. Formally, given an RGB video clip $x _ { i } ^ { V }$ and a text description $x _ { i } ^ { T }$, we feed the video $x _ { i } ^ { V }$ into the video encoder $E _ { V }$ and the text $x _ { i } ^ { T }$ into the text encoder $E _ { T }$ to obtain the feature representation of each sample in a common embedding space by projecting the encoded feature to a common dimension represented by:
\begin{equation}
\label{encoder2}
f^{V} = \psi _ { V } ( E _ { V } ( x _ { i } ^ { V } )) , \quad f^{T} = \psi _ { T } (E _ { T } ( x _ { i } ^ { T } )).
\end{equation}
where $ f^{V} \in \mathbb{R} ^ { C _ { V } }$ and  $ f^{T} \in \mathbb{R} ^ { C _ { T } } $ represent the RGB video and text embedding after the projection layers $\psi _ { V }$ and $\psi _ { T }$, respectively. The dimensions of $C_V$, and $C_T$ are identical after passing through the projection layer.

Within the VG4D framework, we employ X-CLIP~\cite{XCLIP} as both the text and video encoders. X-CLIP builds upon CLIP by incorporating cross-frame attention mechanisms and video-specific hinting techniques. These enhancements enable the extraction of temporal cross-frame interaction information and the generation of instance-level discriminative textual representations, respectively.

\subsection{Cross-Modal Learning}
To learn more transferable representation from VLM, we introduce a cross-modal learning objective to jointly optimize the correlation alignment across language, RGB video, and 4D point cloud. The overall architecture of our method, shown in Fig.~\ref{fig2},
contains language encoder $E _ { T }$, point cloud encoder $E _ { P }$ and visual encoder $E _ { V }$, which respectively embed the three modalities into text feature $f ^ { T } \in \mathbb{R} ^ { C _ { T } }$, image feature $f ^ { V } \in \mathbb{R} ^ { C _ { V } }$ and point cloud feature $f ^ { P } \in \mathbb{R} ^ { C _ { P } }$, where $C$ is the embedding dimension. Through normalization, we constrain the output of each projection network to reside within a unit hypersphere, enabling us to measure feature similarity using the inner product. Our cross-modal contrastive learning mainly jointly optimizes the correlation alignment across languages, images, and point clouds via semantic-level language-4D alignment and instance-level image-4D alignment. 
The core idea of semantic-level language-3D alignment is to drive the feature of 4D action instances and the corresponding action category text closer. In a training mini-batch of size $N$, with $K$ action categories, we calculate the loss function between the language and 4D point cloud as:
\begin{equation}
{L} _{(pc,text)}=\frac{1}{N}\sum _{i \in N} - \log \frac{exp(f_{i}^{T} \cdot f_{i}^{P})}{\sum_{\substack{j \in K}} exp(f_{j}^{T}\cdot f_{i}^{P})}.
\end{equation}
We further introduce the contrastive alignment between instance-wise RGB video and 4D instances. The contrastive aligned objective ${L} _{(pc,video)}$ across point cloud and image is formulated as:

\begin{equation}
{L} _{(pc,video)}=\frac{1}{N}\sum _{i \in N} - \log \frac{exp(f_{i}^{V} \cdot f_{i}^{P})}{\sum_{\substack{j \in N}} exp(f_{j}^{V}\cdot f_{i}^{P})}.
\end{equation}

Finally, we minimize the contrastive loss for all modality pairs with different coefficients $\alpha$ and $\beta$,
\begin{equation}
\label{loss_cl}
L_{cl}= \alpha \textit{L} _{(pc,video)} + \beta {L} _{(pc,text)}.
\end{equation}

During cross-modal learning, We use $L_{cl}$ as supervision to fine-tune the im-PSTNet model that has been pre-trained to learn 4D representations. Note that the VLM is frozen at this stage. Our VG4D also includes two classification heads to classify the 4D features and RGB video features extracted by im-PSTNet and Video encoder, respectively. Our final loss is as follows:
\begin{equation}
\label{loss_final}
L_{final}= L_{cl} + \theta{L}_{pc} +\gamma {L}_{rgb}.
\end{equation}

In the testing phase, we ensemble the im-PSTNet with the VLM. Specifically, we fuse four 4D-text, RGB-text, 4D, and RGB scores as the final classification result.

\subsection{im-PSTNet}
\label{im-PSTNet}
In this subsection, we first briefly review the classical and widely used network in dynamic point cloud modeling. Then, we present how we modernize the classical architecture PSTNet~\cite{fan2021pstnet} into im-PSTNet, the improved version of PSTNet with effective performance. Finally, we detail the network structure of im-PSTNet. As depicted in Fig.\ref{fig2}, the im-PSTNet mainly consists of the spatial extractor and im-PSTConv units. 

\noindent
\textbf{Review of PSTNet}.
%
PSTNet is a spatial and temporal decoupled feature extractor for 4D point clouds. PSTNet uses point spatio-temporal (PST) convolution to extract informative representations of point cloud sequences. The spatio-temporal features are denoted as $h(p_{i}^{t})$, which are aggregated in the following manner:
\begin{equation}
\label{im-PSTConv}
h(p_{i}^{t})=\max_{p_{j}^{t}\in N(p_{i}^{t})}\left\{ \zeta(f_{j}^{t^{\prime}},(x_{j}^{t^{\prime}}-x_{i}^{t}))\right\}.
\end{equation}
where $f_{j}^{t^{\prime}}$ and $x_{j}^{t^{\prime}}$ represent the feature and coordinates of points in the spatio-temporal neighbors' points, and $\zeta$ represents the MLP layers.

Our paper focuses on enhancing the 4D point cloud recognition task by utilizing VLM pre-training knowledge. To achieve this, we propose the im-PSTNet, an improved and modernized version of PSTNet, as the 4D backbone of VG4D.

\noindent
\textbf{From PSTNet to im-PSTNet}.
Our exploration mainly focuses on training strategy modernization and network architecture modernization.
We first employ data augmentation techniques on point cloud video clips. Contrary to PSTNet's method of training and testing all potential clip segments, we employ a data augmentation strategy for frame sampling, significantly reducing both training and testing durations. Specifically, we first divide each point cloud video into \textit{T} segments with equal duration. During the training phase, a frame is randomly sampled from each segment, while in the testing phase, a frame is selected from the middle position, in each segment.
Our experiments show that using the cosine learning rate decay method can lead to better training results than using the step decay method used by PSTNet. As a result, we adopt the cosine learning rate decay method in our im-PSTNet.
In terms of network structure, we use the radius $r$ of the search neighborhood point to normalize $\Delta_{x}=x_{j}^{t}-x_{i}^{t}$, which will make the value of the relative coordinates less small, which is conducive to network optimization.
In addition, to better aggregate the features of the spatio-temporal neighbors, we increase the feature $f_{i}^{t}$ of the center point itself to update the features of each center point.
The spatio-temporal features are aggregated in the following manner:
\begin{equation}
\label{im-PSTConv}
h(p_{i}^{t})=\max_{p_{j}^{t}\in N(p_{i}^{t})}\left\{ \zeta(f_{j}^{t^{\prime}},f_{i}^{t},(x_{j}^{t^{\prime}}-x_{i}^{t})/r)\right\}.
\end{equation}
where $f_{j}^{t^{\prime}}$ and $x_{j}^{t^{\prime}}$ represent the feature and coordinates of points in the spatio-temporal neighbors' points, $r$ represents the radius of searching for spatio-temporal neighbor points, and $\zeta$ represents the MLP layers.

\noindent
\textbf{Architecture of im-PSTNet}.
\textbf{Spatial extractor} is designed to extract the initial features from the N points in each frame, which consists of four sub-modules: point sampling, grouping, MLP layers, and max-pooling. In the point sampling layer, given a spatial subsampling rate $S _ { s }$, the iterative farthest point sampling(FPS) method is used to subsample \textit{N} points to $N ^ { \prime } = [ \frac { N } { s _ { s } } ]$ centroids in each frame. Then the grouping layer searches for a few neighboring points around each centroid to construct a local region for the points subsampled after the FPS. After applying MLP and MAX pooling, the resulting output will contain the coordinates $(x_{i}^{t})$ and features $(p_{i}^{t})$ of each point that has undergone downsampling.
\textbf{im-PSTConv} is improved based on point spatio-temporal (PST) convolution~\cite{fan2021pstnet}, used to extract spatio-temporal information. The difference between im-PSTConv and spatial extractor is that im-PSTConv will group spatio-temporal points by building point pipes. It searches for spatio-temporal neighbors across frames, so this module is called a cross-grouping module.

\section{EXPERIMENTS}

In this section, we describe the implementation details, experiment setup, and experimental results. 

\begin{table*}[!t]
\centering
\captionsetup{justification=centering} 
\caption{Accuracies ($\%$) of different methods on the NTU RGB+D 60 and NTU RGB+D 120 datasets. Best in bold, second-best underlined.}
\begin{tabular*}{\textwidth}{@{\extracolsep{\fill}}c c c *{4}{c}@{}}
\toprule
\multirow{2}*{Method} & \multirow{2}*{Venue} & \multirow{2}*{Modality} & \multicolumn{2}{c}{NTU60} & \multicolumn{2}{c}{NTU120} \\
\cmidrule(lr){4-5} \cmidrule(lr){6-7}
&  &  & Cross-subject & Cross-view & Cross-subject & Cross-setup \\
\midrule[0.3mm]
\multicolumn{7}{c}{Uni-modal recognition methods}\\ \hline
3DV-PointNet++ & CVPR'20~\cite{DBLP:conf/cvpr/YanchengWang} & Point Cloud & 88.8 & 96.3 & 82.4 & 93.5 \\ 
PSTNet & ICLR'21~\cite{fan2021pstnet} & Point Cloud & 90.5 & 96.5 & 87.0 & 93.8 \\
PSTNet++ & TPAMI'21~\cite{DBLP:journals/pami/FanYYK22} & Point Cloud & 91.4 & 96.7 & 88.6 & 93.8 \\
PST-Transformer & TPAMI'22~\cite{DBLP:journals/pami/FanYK23} & Point Cloud & 91.0 & 96.4 & 87.5 & \textbf{\color{blue}94.0} \\
Kinet & CVPR'22~\cite{DBLP:conf/cvpr/ZhongZH0TM22} & Point Cloud & 92.3 & 96.4 & - & - \\
GeometryMotion & TCSVT'21~\cite{tcsv/GMNet} & Point Cloud & 92.7 & 98.9 & 90.1 & 93.6 \\
APSNet & TIP'22~\cite{apsnet9844448} & Point Cloud & 91.5 & 98.2 & 88.3 & 92.5 \\
\hline
\addlinespace[0.5em]
 Ours& - & Point Cloud & \textbf{\color{blue}93.9} & \textbf{\color{blue}98.9} & \textbf{\color{blue}90.3} & 92.0 \\ 
\midrule[0.3mm]
\multicolumn{7}{c}{Multi-modal recognition methods}\\ \hline 
CAPF & CVPR'22~\cite{cafp9879509} & RGB + Depth & 94.2 & 97.3 & - & - \\ 
PA-AWCNN & ICRA'22~\cite{PA-AWCNN} & RGB + Depth & 92.8 & 95.7 & - & - \\ 
Feature Fusion  & IROS'19~\cite{feature-fusiuon} & RGB + Skeleton & 85.4 & 91.6 & - & - \\ 
VPN & ECCV'20~\cite{DBLP:conf/eccv/DasSDBT20} & RGB + Skeleton & 93.5 & 96.2 & 86.3 & 87.8 \\ 
STAR-Transformer & WACV'23~\cite{DBLP:conf/wacv/AhnKHK23} & RGB + Pose & 92.0 & 96.5 & 90.3 & 92.7 \\ 
MMNet & TPAMI'22~\cite{MMnet9782511} & RGB + Pose & 96.0 & 98.8 & 92.9 & 94.4 \\ 
PoseC3D & CVPR'22~\cite{posec3d9879048} & RGB + Pose & \underline{97.0} & \underline{99.6} & \underline{96.4} & \underline{95.4} \\ 
\hline
\addlinespace[0.5em] 
 Ours& - & RGB + Point Cloud & \textbf{\color{blue}97.6} & \textbf{\color{blue}99.8} & \textbf{\color{blue}96.8} & \textbf{\color{blue}97.6} \\ 
\bottomrule
\end{tabular*}
\label{tab:1}
\end{table*}

\noindent
\textbf{Dataset.} NTU RGB+D~\cite{DBLP:conf/cvpr/ShahroudyLNW16} is a large-scale benchmark dataset for action recognition, which contains 56,880 videos collected from 40 subjects performing 60 different actions in 80 camera views. The videos are captured using Kinect V2 to collect four data modalities: RGB frames, depth maps, 3D joint information, and IR sequences. Cross-subject and cross-view evaluations are adopted. NTU RGB+D 120~\cite{Liu_2019_NTURGBD120} is an extension of NTU60, with 120 action classes and 114,480 videos. The action classes include daily actions, health-related actions, and mutual actions. This dataset is also collected with Kinect V2 and shares the same modality with NTU RGB+D 60 dataset. We use the cross-subject and cross-setup evaluation protocols on the NTU RGB+D 120 dataset.

\noindent
\textbf{Implementation Details.} 
For point cloud data preparation, we follow PSTNet to convert depth maps to point cloud sequences, in which we sample 2048 points in each frame. We use the SGD optimizer with cosine learning rate decay for optimization. The initial learning rate, the weight decay, and the batch size are empirically set as 0.01, 0.1, and 32, respectively. We pre-train 120 epochs on NTU RGB+D 60 and NTU RGB+D 120. The number of neighboring points K and the spatial search radius \textit{r} at the grouping module are set as 9 and 0.1, respectively. Following PSTNet, we set the clip length and frame sampling stride to 23 and 2, respectively. 
For the RGB modality, we set the number of input frames to 8, using the same frame sampling method in the point cloud video. 
We use the pre-trained X-CLIP-B/16 model on Kinetics600~\cite{carreira2018short} to fine-tune for 30 epochs on the NTU RGB+D dataset.
In contrastive learning, we train for 30 epochs, the learning rate decays from 0.001 to 0.0001, and the other settings are the same as those of the pre-trained 4D encoder.
All our experiments are performed on two NVIDIA 12G 3080Ti GPUs.
\subsection{Comparison with state-of-the-art methods}
In Table~\ref{tab:1}, we compare our proposed method with other methods on the two datasets. Our im-PSTNet outperforms other single modal baseline methods under most of the settings on both datasets, which demonstrates the effectiveness of our im-PSTNet for 4D action recognition on large-scale datasets. Concurrently, our VG4D achieves state-of-the-art results on multi-modal baseline approaches.
\subsection{Ablation Study}
\noindent
\textbf{Comparison of Different Losses.} We report the effect of using different losses when fine-tuning im-PSTNet in Table~\ref{tabloss}. Among them, the pc-rgb and pc-text loss represent the comparative learning loss of point cloud and RGB video and text, respectively. 
The cls loss represents the cross-entropy loss of im-PSTNet. In particular, after removing the two contrastive learning losses, the accuracy of action recognition dropped significantly, which proves the effectiveness of the contrastive learning method we proposed. 
\noindent
\textbf{Comparison of Different Fusion Methods.} To further show the effectiveness of our method, we compare different combinations of classification scores in Table~\ref{tabscore}.
PC, PC-Text, RGB, and RGB-Text represent the FC classification score of the point cloud, the comparison score of the point cloud and text, the FC classification score of RGB, and the comparison score of RGB and text. CML representation using cross-modal learning in VG4D.

\noindent
\textbf{Comparison of Different Improvements.}
In Table~\ref{tabim-PSTNet}, we report the results of our proposed 4D encoder im-PSTNet when using different modules compared to the original PSTNet. 
As can be seen from the table, the im-PSTNet has a greater improvement in 4D action recognition compared to the PSTNet baseline.

\noindent
\textbf{Comparison of Different VLMs.} We experiment with different VLMs under our framework. 
We report the results of using X-CLIP and Vita-CLIP pre-trained models and the effect of integrating our framework in Table~\ref{tab:VLM}. 
As can be seen from the Table, impressive results can also be achieved using Vita-CLIP. 
This shows that the framework we proposed is universal, and we can integrate VLM with excellent performance into our framework. 

\begin{table}[!t]
    \centering
    \captionsetup{justification=centering} 
    \caption{Cross-subject classification accuracies (\%) of different loss on the NTU RGB+D 120 dataset.}
    \begin{tabular}{@{} >{\centering}p{6cm} c @{}}
    \toprule
    \textbf{Methods} & \textbf{Accuracy (\%)} \\ 
    \midrule
    VG4D & \textbf{96.8} \\ 
    VG4D (w/o cls loss) & 96.0 \\ 
    VG4D (w/o pc-rgb loss) & 95.4 \\ 
    VG4D (w/o pc-text loss) & 95.0 \\ 
    \bottomrule
    \end{tabular}
    \label{tabloss}
\end{table}

\begin{table}[!t]
    \centering
    \caption{Cross-subject classification accuracies (\%) of different categorical score combination on the NTU RGB+D 120 dataset. CML stands for Cross-Modal Learning in VG4D.}
    \begin{tabular}{cccccc}
    \toprule
    CML & PC & PC-Text & RGB-Text & RGB & Accuracy (\%) \\ 
    \midrule
    & $\checkmark$ & & $\checkmark$ & & 96.3 \\ 
    $\checkmark$ & & $\checkmark$ & & &  95.2 \\ 
    $\checkmark$ & $\checkmark$ & & $\checkmark$ & &  96.1 \\ 
    $\checkmark$ & $\checkmark$ & & $\checkmark$ & $\checkmark$ &  96.7 \\ 
    $\checkmark$ & $\checkmark$ & $\checkmark$ & $\checkmark$ & &  96.5 \\ 
    $\checkmark$ & $\checkmark$ & $\checkmark$ & $\checkmark$ & $\checkmark$ & \textbf{96.8} \\ 
    \bottomrule
    \end{tabular}
    \label{tabscore}
\end{table}

\begin{table}[!t]
\centering
\captionsetup{justification=centering} 
\caption{Additive study of sequentially applying training strategies and architecture modernization on NTU RGB+D 120 dataset. We use green and yellow background colors to denote training strategy and model optimization respectively.}
\begin{tabularx}{\linewidth}{>{\centering\arraybackslash}X c c}
\toprule
\textbf{Method} & \textbf{Accuracy (\%)} & \boldmath$\Delta$ \\
\midrule
PSTNet & 88.6 & - \\
\rowcolor{green!10} + Random frame sampling data & 89.0 & +0.4 \\
\rowcolor{green!10} + Step Decay $\rightarrow$ Cosine Decay & 89.2 & +0.2 \\
\rowcolor{yellow!10} + Normalizing $\Delta _p$ (Equation~\ref{im-PSTConv}) & 89.9 & +0.7 \\
\rowcolor{yellow!10} + Feature aggregation (im-PSTNet) & \textbf{90.3} & +0.4 \\
\bottomrule
\end{tabularx}
\label{tabim-PSTNet}
\end{table}

\begin{table}[!t] 
\centering
\caption{Cross-subject classification accuracy (\%) of different vision language on NTU RGB+D 120 dataset.}
\begin{tabular}{>{\centering\arraybackslash}p{3cm}>{\centering\arraybackslash}p{3cm}c}
\toprule
Method & Modality & Accuracy (\%) \\ 
\midrule
X-CLIP & RGB & 95.2 \\
Vita-CLIP & RGB & 95.1 \\
VG4D (Vita-CLIP) & RGB + Point Cloud & 95.5 \\
VG4D (X-CLIP) & RGB + Point Cloud & \textbf{96.8} \\
\bottomrule
\end{tabular}
\label{tab:VLM}
\end{table}



\begin{figure}[!t]
\centering
\includegraphics[width=3.5in]{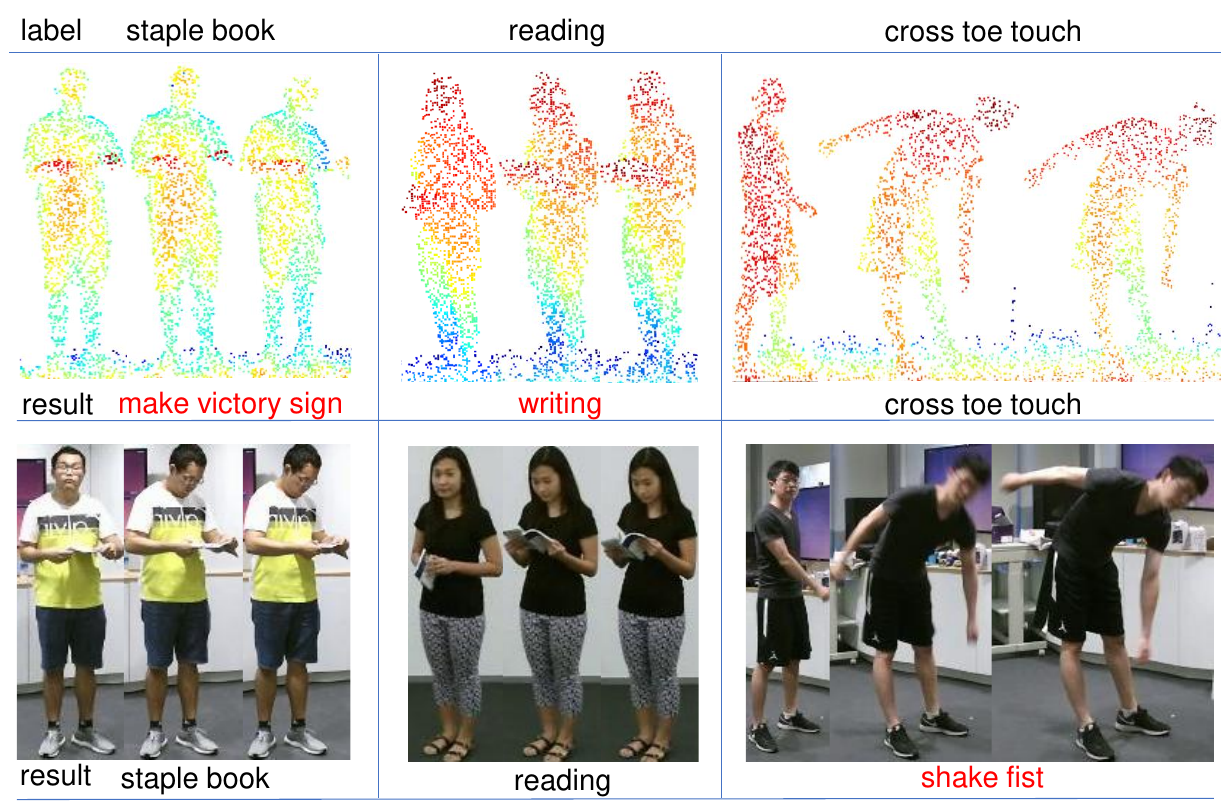}
\vspace{-2em}
\caption{Action classification cases for some different modalities. }
\label{cases}
\vspace{-1em}
\end{figure}
 
\subsection{Further analysis}

\noindent

\noindent
\textbf{Hard Cases.} Some classification failure cases of im-PSTNet are shown in Fig.~\ref{cases}. Recognition failure in point cloud modalities is often caused by the absence of detailed information, such as hand and object movements, which are crucial for distinguishing actions that only involve hand movements. Conversely, RGB mode recognition fails due to the lack of depth information, which is precisely what the point cloud mode provides.
\section{CONCLUSION}
In this paper, we explore how VLM knowledge benefits 4D point cloud understanding. 
We present a novel VLM goes 4D framework with an effective 4D backbone named im-PSTNet to learn better 4D representations. 
To efficiently transfer VLM’s image and text features to a 4D network, we propose a novel cross-modal contrastive learning scheme. 
Our VG4D approach has achieved state-of-the-art performance on various large-scale action recognition datasets. 
Additionally, our proposed im-PSTNet can be utilized as a robust baseline for 4D recognition. 
We hope that this work can inspire action recognition research in the future.





{\small
\bibliographystyle{IEEEtran}
\bibliography{egbib}

\begin{thebibliography}{10}
\providecommand{\url}[1]{#1}
\csname url@rmstyle\endcsname
\providecommand{\newblock}{\relax}
\providecommand{\bibinfo}[2]{#2}
\providecommand\BIBentrySTDinterwordspacing{\spaceskip=0pt\relax}
\providecommand\BIBentryALTinterwordstretchfactor{4}
\providecommand\BIBentryALTinterwordspacing{\spaceskip=\fontdimen2\font plus
\BIBentryALTinterwordstretchfactor\fontdimen3\font minus
  \fontdimen4\font\relax}
\providecommand\BIBforeignlanguage[2]{{%
\expandafter\ifx\csname l@#1\endcsname\relax
\typeout{** WARNING: IEEEtran.bst: No hyphenation pattern has been}%
\typeout{** loaded for the language `#1'. Using the pattern for}%
\typeout{** the default language instead.}%
\else
\language=\csname l@#1\endcsname
\fi
#2}}

\bibitem{ICRApc1}
J.~Liang and A.~Boularias, ``Learning category-level manipulation tasks from
  point clouds with dynamic graph cnns,'' in \emph{ICRA}, 2023.

\bibitem{ICRArgb2}
D.~Seichter, M.~Köhler, B.~Lewandowski, T.~Wengefeld, and H.-M. Gross,
  ``Efficient rgb-d semantic segmentation for indoor scene analysis,'' in
  \emph{ICRA}, 2021.

\bibitem{huang2023visual}
C.~Huang, O.~Mees, A.~Zeng, and W.~Burgard, ``Visual language maps for robot
  navigation,'' in \emph{ICRA}, 2023.

\bibitem{chen2023open}
B.~Chen, F.~Xia, B.~Ichter, K.~Rao, K.~Gopalakrishnan, M.~S. Ryoo, A.~Stone,
  and D.~Kappler, ``Open-vocabulary queryable scene representations for real
  world planning,'' in \emph{ICRA}, 2023.

\bibitem{ral9961854}
D.~Wang and Z.-X. Yang, ``Self-supervised point cloud understanding via mask
  transformer and contrastive learning,'' \emph{RA-L}, 2023.

\bibitem{DBLP:conf/nips/QiYSG17}
C.~R. Qi, L.~Yi, H.~Su, and L.~J. Guibas, ``Pointnet++: Deep hierarchical
  feature learning on point sets in a metric space,'' in \emph{NeurIPS}, 2017.

\bibitem{fang2024explore}
Z.~Fang, X.~Li, X.~Li, J.~M. Buhmann, C.~C. Loy, and M.~Liu, ``Explore
  in-context learning for 3d point cloud understanding,'' \emph{NeurIPS}, 2024.

\bibitem{fang2024modelnet}
Z.~Fang, X.~Li, X.~Li, S.~Zhao, and M.~Liu, ``Modelnet-o: A large-scale
  synthetic dataset for occlusion-aware point cloud classification,''
  \emph{arXiv preprint arXiv:2401.08210}, 2024.

\bibitem{ovsep20204d}
A.~O{\v{s}}ep, P.~Voigtlaender, M.~Weber, J.~Luiten, and B.~Leibe, ``4d generic
  video object proposals,'' in \emph{ICRA}, 2020.

\bibitem{DBLP:journals/pami/FanYK23}
H.~Fan, Y.~Yang, and M.~S. Kankanhalli, ``Point spatio-temporal transformer
  networks for point cloud video modeling,'' \emph{TPAMI}, 2023.

\bibitem{Zeng_2023_CVPR}
Y.~Zeng, C.~Jiang, J.~Mao, J.~Han, C.~Ye, Q.~Huang, D.-Y. Yeung, Z.~Yang,
  X.~Liang, and H.~Xu, ``Clip2: Contrastive language-image-point pretraining
  from real-world point cloud data,'' in \emph{CVPR}, 2023.

\bibitem{hegde2023clip}
D.~Hegde, J.~M.~J. Valanarasu, and V.~M. Patel, ``Clip goes 3d: Leveraging
  prompt tuning for language grounded 3d recognition,'' \emph{arXiv preprint
  arXiv:2303.11313}, 2023.

\bibitem{hanoonavificlip}
H.~Rasheed, M.~U. khattak, M.~Maaz, S.~Khan, and F.~S. Khan, ``Finetuned clip
  models are efficient video learners,'' in \emph{CVPR}, 2023.

\bibitem{wu2023open}
J.~Wu, X.~Li, S.~Xu, H.~Yuan, H.~Ding, Y.~Yang, X.~Li, J.~Zhang, Y.~Tong,
  X.~Jiang, B.~Ghanem, and D.~Tao, ``Towards open vocabulary learning: A
  survey,'' \emph{T-PAMI}, 2024.

\bibitem{xu2023dst}
S.~Xu, X.~Li, S.~Wu, W.~Zhang, Y.~Li, G.~Cheng, Y.~Tong, K.~Chen, and C.~C.
  Loy, ``Dst-det: Simple dynamic self-training for open-vocabulary object
  detection,'' \emph{arXiv preprint arXiv:2310.01393}, 2023.

\bibitem{Xue_2023_CVPR}
L.~Xue, M.~Gao, C.~Xing, R.~Mart{\'\i}n-Mart{\'\i}n, J.~Wu, C.~Xiong, R.~Xu,
  J.~C. Niebles, and S.~Savarese, ``Ulip: Learning a unified representation of
  language, images, and point clouds for 3d understanding,'' in \emph{CVPR},
  2023.

\bibitem{XCLIP}
B.~Ni, H.~Peng, M.~Chen, S.~Zhang, G.~Meng, J.~Fu, S.~Xiang, and H.~Ling,
  ``Expanding language-image pretrained models for general video recognition,''
  in \emph{ECCV}, 2022.

\bibitem{ral9956014}
J.~Chen, Y.~Zhang, F.~Ma, and Z.~Tan, ``Eb-lg module for 3d point cloud
  classification and segmentation,'' \emph{RA-L}, 2023.

\bibitem{DBLP:conf/cvpr/QiSMG17}
C.~R. Qi, H.~Su, K.~Mo, and L.~J. Guibas, ``Pointnet: Deep learning on point
  sets for 3d classification and segmentation,'' in \emph{CVPR}, 2017.

\bibitem{qian2022pointnext}
G.~Qian, Y.~Li, H.~Peng, J.~Mai, H.~Hammoud, M.~Elhoseiny, and B.~Ghanem,
  ``Pointnext: Revisiting pointnet++ with improved training and scaling
  strategies,'' in \emph{NeurIPS}, 2022.

\bibitem{dgcnn}
Y.~Wang, Y.~Sun, Z.~Liu, S.~E. Sarma, M.~M. Bronstein, and J.~M. Solomon,
  ``Dynamic graph cnn for learning on point clouds,'' \emph{TOG}, 2019.

\bibitem{DBLP:conf/iccv/ZhaoJJTK21}
H.~Zhao, L.~Jiang, J.~Jia, P.~H.~S. Torr, and V.~Koltun, ``Point transformer,''
  in \emph{ICCV}, 2021.

\bibitem{DBLP:journals/ral/ChenWKYR23}
L.~Chen, H.~Wang, H.~Kong, W.~Yang, and M.~Ren, ``Ptc-net: Point-wise
  transformer with sparse convolution network for place recognition,''
  \emph{RA-L}, 2023.

\bibitem{DBLP:conf/cvpr/YuTR00L22}
X.~Yu, L.~Tang, Y.~Rao, T.~Huang, J.~Zhou, and J.~Lu, ``Point-bert:
  Pre-training 3d point cloud transformers with masked point modeling,'' in
  \emph{CVPR}, 2022.

\bibitem{DBLP:conf/eccv/PangWTLTY22}
Y.~Pang, W.~Wang, F.~E.~H. Tay, W.~Liu, Y.~Tian, and L.~Yuan, ``Masked
  autoencoders for point cloud self-supervised learning,'' in \emph{ECCV},
  2022.

\bibitem{yan2022let}
X.~Yan, H.~Zhan, C.~Zheng, J.~Gao, R.~Zhang, S.~Cui, and Z.~Li, ``Let images
  give you more: Point cloud cross-modal training for shape analysis,'' in
  \emph{NeurIPS}, 2022.

\bibitem{fang2023explore}
Z.~Fang, X.~Li, X.~Li, J.~M. Buhmann, C.~C. Loy, and M.~Liu, ``Explore
  in-context learning for 3d point cloud understanding,'' \emph{NeurIPS}, 2023.

\bibitem{DBLP:conf/iccv/LiuYB19}
X.~Liu, M.~Yan, and J.~Bohg, ``Meteornet: Deep learning on dynamic 3d point
  cloud sequences,'' in \emph{ICCV}, 2019.

\bibitem{fan2021pstnet}
H.~Fan, X.~Yu, Y.~Ding, Y.~Yang, and M.~Kankanhalli, ``Pstnet: Point
  spatio-temporal convolution on point cloud sequences,'' in \emph{ICLR}, 2021.

\bibitem{DBLP:journals/pami/FanYYK22}
H.~Fan, X.~Yu, Y.~Yang, and M.~S. Kankanhalli, ``Deep hierarchical
  representation of point cloud videos via spatio-temporal decomposition,''
  \emph{TPAMI}, 2022.

\bibitem{DBLP:conf/cvpr/YanchengWang}
Y.~Wang, Y.~Xiao, F.~Xiong, W.~Jiang, Z.~Cao, J.~T. Zhou, and J.~Yuan, ``3dv:
  3d dynamic voxel for action recognition in depth video,'' in \emph{CVPR},
  2020.

\bibitem{DBLP:conf/cvpr/ZhongZH0TM22}
J.~Zhong, K.~Zhou, Q.~Hu, B.~Wang, N.~Trigoni, and A.~Markham, ``No pain, big
  gain: Classify dynamic point cloud sequences with static models by fitting
  feature-level space-time surfaces,'' in \emph{CVPR}, 2022.

\bibitem{tcsv/GMNet}
J.~Liu and D.~Xu, ``Geometrymotion-net: A strong two-stream baseline for 3d
  action recognition,'' \emph{TCSVT}, 2021.

\bibitem{tmm/GMTNet}
J.~Liu, J.~Guo, and D.~Xu, ``Geometrymotion-transformer: An end-to-end
  framework for 3d action recognition,'' \emph{TMM}, 2022.

\bibitem{wang2024gcnext}
X.~Wang, Q.~Cui, C.~Chen, and M.~Liu, ``Gcnext: Towards the unity of graph
  convolutions for human motion prediction,'' in \emph{AAAI}, 2024.

\bibitem{wang2023skeleton}
X.~Wang, Z.~Fang, X.~Li, X.~Li, C.~Chen, and M.~Liu, ``Skeleton-in-context:
  Unified skeleton sequence modeling with in-context learning,'' \emph{CVPR},
  2024.

\bibitem{wang2023dynamic}
X.~Wang, W.~Zhang, C.~Wang, Y.~Gao, and M.~Liu, ``Dynamic dense graph
  convolutional network for skeleton-based human motion prediction,''
  \emph{TIP}, 2023.

\bibitem{fan21p4transformer}
H.~Fan, Y.~Yang, and M.~Kankanhalli, ``Point 4d transformer networks for
  spatio-temporal modeling in point cloud videos,'' in \emph{CVPR}, 2021.

\bibitem{DBLP:conf/eccv/WenLHDY22}
H.~Wen, Y.~Liu, J.~Huang, B.~Duan, and L.~Yi, ``Point primitive transformer for
  long-term 4d point cloud video understanding,'' in \emph{ECCV}, 2022.

\bibitem{DBLP:conf/wacv/WeiLXKG22}
Y.~Wei, H.~Liu, T.~Xie, Q.~Ke, and Y.~Guo, ``Spatial-temporal transformer for
  3d point cloud sequences,'' in \emph{{WACV}}, 2022.

\bibitem{DBLP:conf/mm/ChenLL0H022}
X.~Chen, W.~Liu, X.~Liu, Y.~Zhang, J.~Han, and T.~Mei, ``{MAPLE:} masked
  pseudo-labeling autoencoder for semi-supervised point cloud action
  recognition,'' in \emph{ACM MM}, 2022.

\bibitem{Shen_2023_CVPR}
Z.~Shen, X.~Sheng, L.~Wang, Y.~Guo, Q.~Liu, and Z.~Xi, ``Pointcmp: Contrastive
  mask prediction for self-supervised learning on point cloud videos,'' in
  \emph{CVPR}, 2023.

\bibitem{huang2022clip2point}
T.~Huang, B.~Dong, Y.~Yang, X.~Huang, R.~W. Lau, W.~Ouyang, and W.~Zuo,
  ``Clip2point: Transfer clip to point cloud classification with image-depth
  pre-training,'' in \emph{ICCV}, 2023.

\bibitem{yang2023aim}
T.~Yang, Y.~Zhu, Y.~Xie, A.~Zhang, C.~Chen, and M.~Li, ``Aim: Adapting image
  models for efficient video understanding,'' in \emph{ICLR}, 2023.

\bibitem{bike}
W.~Wu, X.~Wang, H.~Luo, J.~Wang, Y.~Yang, and W.~Ouyang, ``Bidirectional
  cross-modal knowledge exploration for video recognition with pre-trained
  vision-language models,'' in \emph{CVPR}, 2023.

\bibitem{wasim2023vitaclip}
S.~T. Wasim, M.~Naseer, S.~Khan, F.~S. Khan, and M.~Shah, ``Vita-clip: Video
  and text adaptive clip via multimodal prompting,'' in \emph{CVPR}, 2023.

\bibitem{DBLP:conf/cvpr/ZhangGZLM0QG022}
R.~Zhang, Z.~Guo, W.~Zhang, K.~Li, X.~Miao, B.~Cui, Y.~Qiao, P.~Gao, and H.~Li,
  ``Pointclip: Point cloud understanding by {CLIP},'' in \emph{CVPR}, 2022.

\bibitem{apsnet9844448}
J.~Liu, J.~Guo, and D.~Xu, ``Apsnet: Toward adaptive point sampling for
  efficient 3d action recognition,'' \emph{TIP}, 2022.

\bibitem{cafp9879509}
B.~Zhou, P.~Wang, J.~Wan, Y.~Liang, F.~Wang, D.~Zhang, Z.~Lei, H.~Li, and
  R.~Jin, ``Decoupling and recoupling spatiotemporal representation for
  rgb-d-based motion recognition,'' in \emph{CVPR}, 2022.

\bibitem{PA-AWCNN}
L.~Yao, S.~Liu, C.~Li, S.~Zou, S.~Chen, and D.~Guan, ``Pa-awcnn: Two-stream
  parallel attention adaptive weight network for rgb-d action recognition,'' in
  \emph{ICRA}, 2022.

\bibitem{feature-fusiuon}
G.~Liu, J.~Qian, F.~Wen, X.~Zhu, R.~Ying, and P.~Liu, ``Action recognition
  based on 3d skeleton and rgb frame fusion,'' in \emph{IROS}, 2019.

\bibitem{DBLP:conf/eccv/DasSDBT20}
S.~Das, S.~Sharma, R.~Dai, F.~Br{\'{e}}mond, and M.~Thonnat, ``{VPN:} learning
  video-pose embedding for activities of daily living,'' in \emph{ECCV}, 2020.

\bibitem{DBLP:conf/wacv/AhnKHK23}
D.~Ahn, S.~Kim, H.~Hong, and B.~Ko, ``Star-transformer: {A} spatio-temporal
  cross attention transformer for human action recognition,'' in \emph{WACV},
  2023.

\bibitem{MMnet9782511}
B.~X. Yu, Y.~Liu, X.~Zhang, S.-h. Zhong, and K.~C. Chan, ``Mmnet: A model-based
  multimodal network for human action recognition in rgb-d videos,''
  \emph{TPAMI}, 2023.

\bibitem{posec3d9879048}
H.~Duan, Y.~Zhao, K.~Chen, D.~Lin, and B.~Dai, ``Revisiting skeleton-based
  action recognition,'' in \emph{CVPR}, 2022.

\bibitem{DBLP:conf/cvpr/ShahroudyLNW16}
A.~Shahroudy, J.~Liu, T.~Ng, and G.~Wang, ``{NTU} {RGB+D:} {A} large scale
  dataset for 3d human activity analysis,'' in \emph{CVPR}, 2016.

\bibitem{Liu_2019_NTURGBD120}
J.~Liu, A.~Shahroudy, M.~Perez, G.~Wang, L.-Y. Duan, and A.~C. Kot, ``Ntu rgb+d
  120: A large-scale benchmark for 3d human activity understanding,''
  \emph{TPAMI}, 2019.

\bibitem{carreira2018short}
J.~Carreira, E.~Noland, A.~Banki-Horvath, C.~Hillier, and A.~Zisserman, ``A
  short note about kinetics-600,'' \emph{arXiv preprint arXiv:1808.01340},
  2018.

\end{thebibliography}
}

\end{document}